\documentclass[conference]{IEEEtran}
\IEEEoverridecommandlockouts
\usepackage{cite}
\usepackage{amsmath,amssymb,amsfonts}
\usepackage{algorithmic}
\usepackage{graphicx}
\usepackage{textcomp}
\usepackage{xcolor}
\usepackage[colorlinks=true]{hyperref}
\usepackage{color}
\usepackage{booktabs}
\usepackage{multirow}
\usepackage{movie15}
\usepackage{subcaption}
\usepackage{enumitem}
\usepackage[misc,geometry]{ifsym}
\newcommand{\FigSpect}[1]{
  \begin{subfigure}{0.23\textwidth}
  \includemovie[poster=images/#1.png,mouse,repeat]{0.95\linewidth}{0.7459807073954984\linewidth}{audios/#1.mp3}
  \label{fig:#1_sample}
\end{subfigure}}

\def\BibTeX{{\rm B\kern-.05em{\sc i\kern-.025em b}\kern-.08em
    T\kern-.1667em\lower.7ex\hbox{E}\kern-.125emX}}
\begin{document}

\title{Efficient Multi-Model Fusion with Adversarial Complementary Representation Learning}

% \author{\IEEEauthorblockN{1\textsuperscript{st} Zuheng Kang$^{\dagger}$\textsuperscript{\Letter}\thanks{\textsuperscript{\Letter} Corresponding author: Zuheng Kang, bob.kang@qq.com}}
%     \IEEEauthorblockA{\textit{Ping An Technology (Shenzhen) Co., Ltd.}\\
%         Shenzhen, China \\
%         kangzuheng896@pingan.com.cn}
%     \and
%     \IEEEauthorblockN{2\textsuperscript{nd} Yayun He$^\dagger$\thanks{$^\dagger$ Equal contributions}}
%     \IEEEauthorblockA{\textit{Ping An Technology (Shenzhen) Co., Ltd.}\\
%         Shenzhen, China \\
%         heyayun097@pingan.com.cn}
%     \and
%     \IEEEauthorblockN{3\textsuperscript{rd} Jianzong Wang}
%     \IEEEauthorblockA{\textit{Ping An Technology (Shenzhen) Co., Ltd.}\\
%         Shenzhen, China \\
%         jzwang@188.com}
%     \and
%     \IEEEauthorblockN{4\textsuperscript{th} Junqing Peng}
%     \IEEEauthorblockA{\textit{Ping An Technology (Shenzhen) Co., Ltd.}\\
%         Shenzhen, China \\
%         pengjq@pingan.com.cn}
%     \and
%     \IEEEauthorblockN{5\textsuperscript{th} Jing Xiao}
%     \IEEEauthorblockA{\textit{Ping An Insurance (Group) Company}\\
%         Shenzhen, China \\
%         xiaojing661@pingan.com.cn}
% }

\author{\IEEEauthorblockN{Zuheng Kang$^\dagger$\textsuperscript{\Letter}\thanks{\textsuperscript{\Letter} Corresponding author: Zuheng Kang, bob.kang@qq.com}, Yayun He$^\dagger$\thanks{$^\dagger$ Both authors have equal contributions.}, Jianzong Wang, Junqing Peng, Jing Xiao}
    \IEEEauthorblockA{\textit{Ping An Technology (Shenzhen) Co., Ltd., Shenzhen, China.}}}

% \author{\IEEEauthorblockN{1\textsuperscript{st} Blind Name}
% \IEEEauthorblockA{\textit{Blind Organization} \\
% Blind Address \\
% Blind Email}
% \and
% \IEEEauthorblockN{2\textsuperscript{nd} Blind Name}
% \IEEEauthorblockA{\textit{Blind Organization} \\
% Blind Address \\
% Blind Email}
% \and
% \IEEEauthorblockN{3\textsuperscript{rd} Blind Name}
% \IEEEauthorblockA{\textit{Blind Organization} \\
% Blind Address \\
% Blind Email}
% \and
% \IEEEauthorblockN{4\textsuperscript{th} Blind Name}
% \IEEEauthorblockA{\textit{Blind Organization} \\
% Blind Address \\
% Blind Email}
% \and
% \IEEEauthorblockN{5\textsuperscript{th} Blind Name}
% \IEEEauthorblockA{\textit{Blind Organization} \\
% Blind Address \\
% Blind Email}
% }
\maketitle

\begin{abstract}

    Single-model systems often suffer from deficiencies in tasks such as speaker verification (SV) and image classification, relying heavily on partial prior knowledge during decision-making, resulting in suboptimal performance.
    Although multi-model fusion (MMF) can mitigate some of these issues, redundancy in learned representations may limits improvements.
    To this end, we propose an adversarial complementary representation learning (ACoRL) framework that enables newly trained models to avoid previously acquired knowledge, allowing each individual component model to learn maximally distinct, complementary representations.
    We make three detailed explanations of why this works and experimental results demonstrate that our method more efficiently improves performance compared to traditional MMF.
    Furthermore, attribution analysis validates the model trained under ACoRL acquires more complementary knowledge, highlighting the efficacy of our approach in enhancing efficiency and robustness across tasks.

\end{abstract}


\vspace{0.5em}
\section{Introduction}
\label{sec:intro}

Multi-model fusion (MMF) has demonstrated great potential
to achieve superior overall performance compared to individual, as distinct component models may contain complementary capabilities to avoid their limitations.

Despite various applications and tasks using quite different architectures and processing methods, there are commonalities exist in the core logic -- MMF for the model exists throughout the model inference pipeline, including input data$ _a $, early, mid, and late stages of the model$ _b $, and final inference output$ _c $ \cite{huang2020fusion}.
(a) Data-level fusion can merge datasets with fully or partially overlapping labels at the label level to incorporate more training data \cite{Kang2022SpeechEQSE}.
Additionally, multi-task learning enables joint training on datasets with disparate labels using different optimization objectives \cite{Kang2022SpeechEQSE}.
Data augmentation via generative models like generative adversarial networks (GANs) also enables the fusion of real and synthetic data \cite{Lin2022ANM}.
(b) Model-level fusion utilizes intermediate representations for fusion.
Early fusion leverages multiple engineered features to allow models to fully exploit information in the data \cite{Gao2019MultimodelFM,peng2021effective,xu2020social,polinati2019review,lin2022multi,tian2021multi}.
Mid-term fusion means that each type of data is first processed through its own network and then fused at some intermediate modeling layer.
Late fusion is more flexible to aggregate high-level features of individually trained models optimized under different conditions.
Varying training induces slightly different model specializations, so integrating them may improve overall performance \cite{Lin2022ANM,Li2019AdaptingIS,deb2022multi,gao2019multi}.
These models are typically fused by concatenation, addition, multiplication, or attention mechanism.
(c) Output-level fusion commonly integrates the predictions of multiple pre-trained models to achieve improved performance \cite{Deng2019DeepMF,Jiang2020ATMFNAM,zhang2020feature}.
This technique is frequently employed in various competitions.
For example, in speaker verification (SV) challenges, fusion of scoring results is often used in the optimization pipeline \cite{huh2023voxsrc,sadjadi20222021,zhao2021speakin}, deepfake detection \cite{yi2023audio}, and many other challenges \cite{Liao2020EarlyBO}.
However, although MMF is effective in improving overall system performance, it has some limitations.
The constituent models used in MMF are often very similar in nature, which limits the effectiveness of model integration.
Also, integrating too many models significantly increases computational costs, making it infeasible in resource-constrained environments.

There are a number of similar methods that attempt to train a new model that avoid previously learned knowledge, but they vary in their approach, and purpose.
Shen et al. \cite{Shen2018MEALME} proposed an adversarial learning based model distillation method, which aims to incorporate the student model to learn the knowledge of the teacher's network, but not to learn its similar knowledge from the corresponding layers in the teacher's model.
However, the problem with this framework is that the student model is only learning the teacher's network, and the incorrect knowledge learned from the teacher's network is not modified by correct labeling, which may lead to the magnifying the incorrect knowledge.
Nam et al. \cite{Nam2021DiversityMW} attempted to utilize the perturbation method so that the student model would absorb as much knowledge as possible from the individual teacher models, thereby distilling a single model that would perform as a complete set.
Although this work distills a model of learning knowledge that is as different as possible from the teacher's knowledge, the model's performance is only close to that of the previous ensembled model, but does not surpass it.
Zhang et al. \cite{Zhang2018AdversarialCL}, most similar to our idea, proposed an adversarial based object localization framework that avoid area that the previous models had learned.
However, this method only apply to the object detection task, and it cannot apply to other tasks.
Most similar to our idea, Zhang et al. \cite{Zhang2018AdversarialCL} proposed an adversarial-based object localization framework which avoids regions learned by previous models.
However, this approach is only applicable to the object localization task and can not be generalized, and cannot apply to the models more than two.
There are many other researches about ensemble learning that are meaningful to the form of our idea \cite{cao2020ensemble,ganaie2022ensemble,dong2020survey,mohammed2023comprehensive,zhou2023inverse}.

\begin{figure*}[t]
    \centering
    \includegraphics[width=0.95\textwidth]{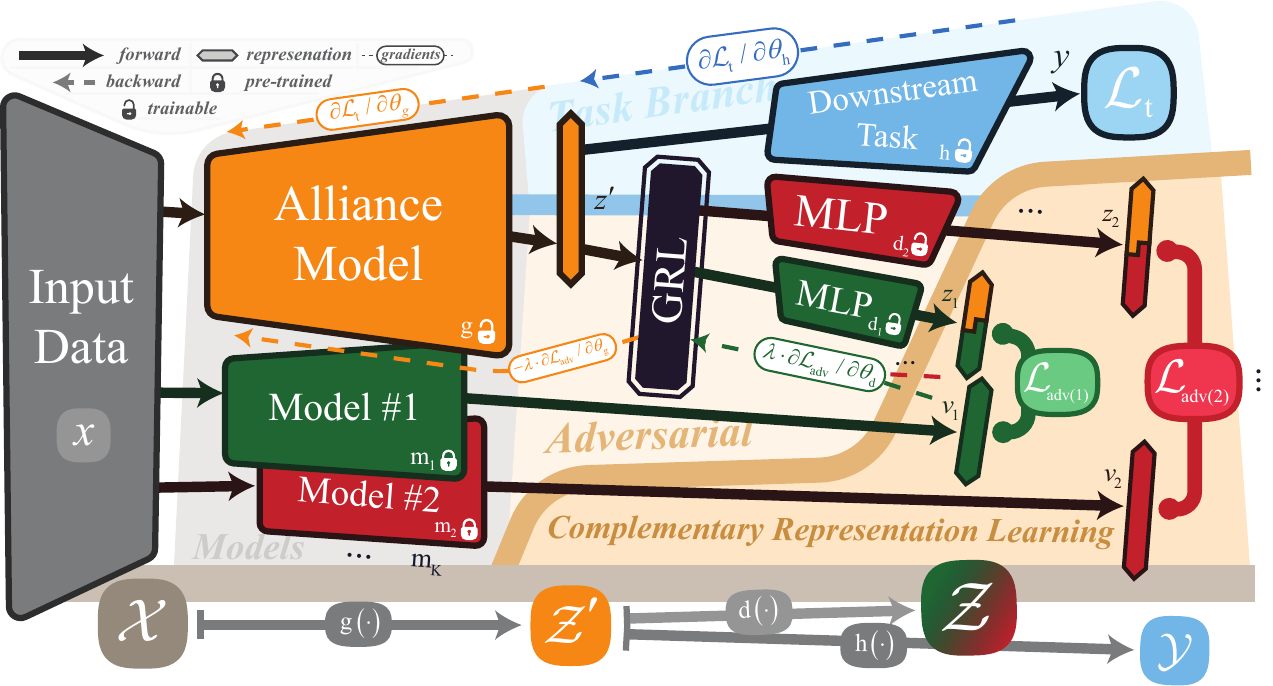}
    \caption{The overview of ACoRL framework.}
    \label{fig:acorl}
    % \vspace{0.5em}
\end{figure*}

Also, in machine learning, bagging and boosting are popular ensemble methods that improve system performance.
Bagging trains multiple models on different bootstrap samples of the training data to reduce variance.
Boosting sequentially fits models to emphasize previously misclassified instances, thereby reducing bias.
However, random sampling of bagging data still does not solve the training redundancy problem.
Boosting uses a sequential approach that limits parallelization and reduces efficiency.
Additionally, some tasks are not amenable to boosting's error-focused learning.

To fully address the challenges of MMF, the analytical lens cannot be narrowly confined.
Let's start with a story:
When a person or group seeks allies to maximize their benefits, they tend to ally with someone who is as different as possible from themselves in terms of knowledge, vision, and ability.
However, if they are free to acquire knowledge from a single source, their acquired knowledge tends to closely resemble that of others.
Allying among them may risk substantial redundancy.
Furthermore, too many models require much more computational resources.

\vspace{0.5em}
\noindent
Therefore, we have made the following contributions:
\begin{itemize}
    \item We propose an adversarial complementary representation learning (ACoRL) framework that promotes diversity during multi-model fusion by enabling models to avoid previously acquired knowledge and learn distinct representations.
    \item We theoretically prove and explain how ACoRL can improve the performance of multi-model fusion (MMF) by extending the range of representations in the latent space.
    \item Experimental results and attribution analysis validate that ACoRL can leverage more complementary knowledge, strengthening its ability to improve model performance across tasks.
\end{itemize}

\section{Methodology}
\label{sec:method}

\subsection{The Need for Complementary Representations}

In supervised learning frameworks, source data is generally transformed by the model into a progression of shallow to deep latent representations, which are eventually mapped to downstream tasks.
For a specific task, when training a single model multiple times, variance in model performance is primarily attributable to data quality and differences in the models.
However, for the same data, since the training of a single model is heavily constrained by the optimization objective, newly trained models tend to focus on the most readily learnable and salient aspects of the training data, while neglecting other potentially useful information.
This likely results in a high degree of consistency in the knowledge acquired across models.
In other words, representations are highly similar from the model perspective.
This phenomenon has been widely observed by many other researchers \cite{Shen2018MEALME,Nam2021DiversityMW,Zhang2018AdversarialCL}.
Therefore, it is necessary to ensure that these representations are as complementary as possible so that the individual models can extract the potential knowledge from the training data as comprehensively as possible.
Meanwhile, downstream tasks can still be accomplished with these alternative representations.

\subsection{Adversarial Complementary Representation Learning}

To obtain efficient fusion system capability, we need to supplement the newly allied models with the strengths and weaknesses of the existing models.
To this end, we propose an adversarial complementary representation learning (ACoRL; alias Alliance Learning), which allows the newly allied model to avoid learning previously acquired knowledge as much as possible.
Note:
(1) \textbf{alliance learning} in this paper is the same as adversarial complementary representation learning (ACoRL) framework.
(2) \textbf{alliance model} is the newly trained model that avoid previously learned knowledge.
Seen from Figure \ref{fig:acorl} upper part, the ACoRL framework consists of 2 main branches: the branch with $ K $ pre-trained models $ \mathrm{m}_k\left(\cdot\right) $ (discussed in $ \S $ \ref{sec:pretrained}), where $ k $ is the index, and the branch with trainable alliance model $ \mathrm{g}\left(\cdot\right) $ with model parameters $ \theta_\mathrm{g} $ (in $ \S $ \ref{sec:alliance}).

\vspace{0.5em}
\subsubsection*{Pre-trained Models Knowledge Avoidance}
\label{sec:pretrained}

The input data $ x $ is first transformed into a set of intermediate representations $ v_k $ by its corresponding pre-trained models $ \mathrm{m}_k\left(\cdot\right) $, shown as the green and red branches in Figure \ref{fig:acorl}.
These representations $ v_k $ will be used to represent the knowledge known by the corresponding pre-trained model $ \mathrm{m}_k $ to avoid learning by the newly allied model.

\vspace{0.5em}
\subsubsection*{Alliance Model Training with ACoRL}
\label{sec:alliance}

The input data $ x $ is then converted to the representations $ z'$ via the alliance model $ \mathrm{g}\left(\cdot\right) $, as shown in the orange branch in Figure \ref{fig:acorl}.
Then, $ z' $ is sent to 2 sub-branches:

\vspace{0.2em}
\begin{enumerate}[label=(\roman*)]
    \item \textbf{Task branch}: $ z' $ accomplishes the task training through the downstream task model $ \mathrm{h}\left(\cdot\right) $ with parameters $ \theta_\mathrm{h} $, as shown in the light blue branch.
    The task branch loss and its function are denoted as $ \mathcal{L}_{\mathrm{t}} $ and $ \ell_\mathrm{t} $, in Equation \ref{eq:loss_t}, which depends on the specific task, where $ y $ is the label.
    \begin{equation}
        \mathcal{L} _{\mathrm{t}} =\ell _{\mathrm{t}}\left(\mathrm{h} \circ \mathrm{g}\left( x \right), y \right).
        \label{eq:loss_t}
    \end{equation}
    \item \textbf{ACoRL branch}: As shown in light brown background area in Figure \ref{fig:acorl}. To compare between the representations of the alliance model $ z' $ and the pre-trained model $ v_k $, they should be dimensionally consistent.
    For this, the projection model $ \mathrm{d}_k\left(\cdot\right) $ implemented as a multi-layer perceptron (MLP) with parameters $ \theta_{\mathrm{d}_k} $ is used to project $ z' $ onto $ z_k $ with the same dimensions as $ v_k $.
    To ensure that the alliance model avoids learning the knowledge in each pre-trained model, an adversarial approach is well-suited.
    Specifically, the projection model aims to ensure the alliance model representation $ z_k $ to match the pre-trained model representation $ v_k $ as similarly as possible.
    In contrast, the alliance model will produce representations that are as complementary as possible.
    To achieve this, the gradient reversal layer (GRL) \cite{Ganin2014UnsupervisedDA} can invert the gradient, which enables the projection model to maximize knowledge similarity with the pre-trained model, while the alliance model minimizes it.
    The ACoRL branch loss and its function corresponding to model $ \mathrm{m}_k $ are denoted as $ \mathcal{L}_{\mathrm{adv}\left(k\right)} $ and $ \ell _{\mathrm{adv}\left(k\right)} $, in Equation \ref{eq:loss_adv}.
    Where $ \circ $ denotes the mapping function.
    \begin{equation}
        \mathcal{L} _{\mathrm{adv}\left( k \right)} =\ell _{\mathrm{adv}\left( k \right)}\left(\mathrm{d}_k \circ \mathrm{g} \left( x \right),\mathrm{m}_k\left( x \right) \right).
        \label{eq:loss_adv}
    \end{equation}
\end{enumerate}

\subsubsection*{Overall Loss}

The hyper-parameter $ \lambda $ is introduced to balance the training of the two task branches, and then the overall loss $ \mathcal{L} $ is shown in Equation \ref{eq:loss}.
In this way, the alliance model receives both the gradient of the task branch $ \partial \mathcal{L} _{\mathrm{t}}/\partial \theta _{\mathrm{g}} $ for task learning, and the reversed gradient of ACoRL branch $ -\lambda \cdot \partial \mathcal{L} _{\mathrm{adv}}/\partial \theta _{\mathrm{g}} $ for complementary knowledge acquiring.
The downstream task model recieves the gradient $ \partial \mathcal{L} _{\mathrm{t}}/\partial \theta _{\mathrm{h}} $ for task training.
The projection models recieve the gradient $ \lambda \cdot \partial \mathcal{L} _{\mathrm{adv}}/\partial \theta _{\mathrm{d}} $ for representation learning, shown in Figure \ref{fig:acorl}.
\begin{equation}
    \mathcal{L} =\max_{\theta _{\mathrm{d}}} \min_{\theta _{\mathrm{g}},\theta _{\mathrm{h}}} \left( \mathcal{L} _{\mathrm{t}} -\frac{\lambda}{K}\sum_{k=1}^K{\mathcal{L} _{\mathrm{adv}\left( k \right)}} \right).
    \label{eq:loss}
\end{equation}

\begin{figure*}[t]
    \centering
    \includegraphics[width=0.95\textwidth]{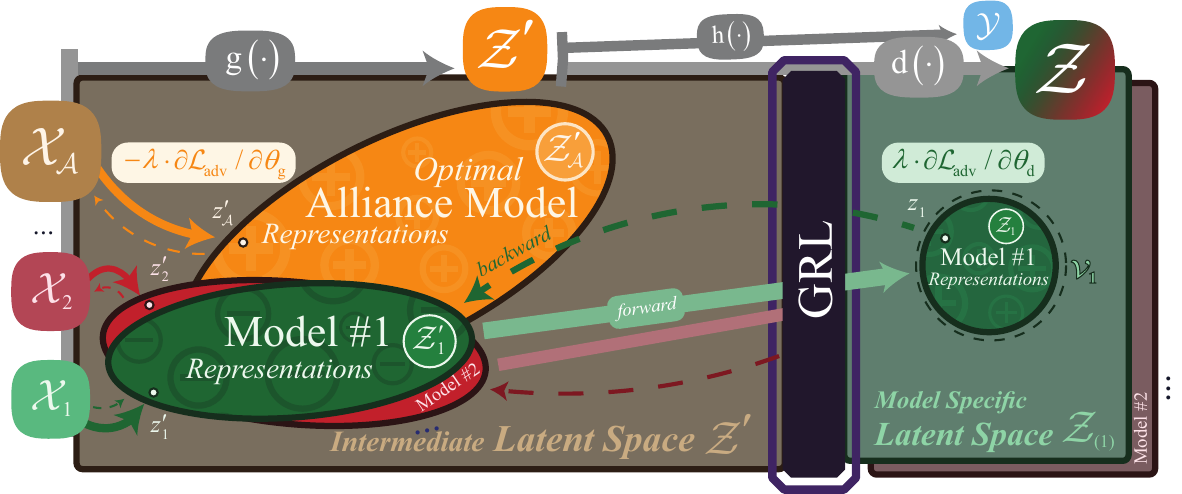}
    \caption{Illustration of the process of Explain 3.}
    \label{fig:explain}
    % \vspace{0.5em}
\end{figure*}

\subsection{Motivation for ACoRL and Explanations}

In order to justify our approach, we identified three explanations, which are the main motivations for proposing the ACoRL framework.
In the end, we proved experimentally that our proposed ACoRL are correct.

\vspace{0.5em}
\noindent
\textbf{Explain 1:}
\textbf{Fundamentally, the mechanism of multi-modal and multi-model fusion for improving the system performance are intrinsically the same.}
\vspace{0.5em}

Multimodality leverages naturally complementary information from distinct physical modalities, such as audio and visual, to improve system performance.
In contrast, unimodal MMF jointly optimizes multiple models from the same data source to learn slightly different but complementary knowledge, thus making the system more robust.
Both approaches attempt to learn a more comprehensive knowledge to improve the overall performance of the system.
However, knowledge learned by multi-model is more distinct than that learned from multi-modal.

\vspace{0.5em}
\noindent
\textbf{Explain 2:}
\textbf{If the sample size is large enough, models that cover a larger and more complete latent representation will have better performance.}
\vspace{0.5em}

According to Explain 1, the explain of multi-modal fusion also applies to MMF.
Yu H. et al. \cite{Huang2021WhatMM} theoretically proved that with sufficient data, the use of more modalities can effectively improve the quality of the latent representation by covering larger latent space, thereby improving the overall system performance.
Similarly, it can be hypothesized that the MMF system could perform better if it is able to obtain a more expansive and comprehensive representation in the latent space.

\vspace{0.5em}
\noindent
\textbf{Explain 3:}
\textbf{ACoRL framework has the potential for larger, higher quality latent representations.}
\vspace{0.5em}

The upper part of Figure \ref{fig:explain} and the lower part of Figure \ref{fig:acorl} are the same.
The large background rectangle represents the ideal maximum boundary of the latent space for accomplishing a given task.
There are three spaces, the input space $ \mathcal{X} $, intermediate latent space $ \mathcal{Z}' $ and model specific latent space $ \mathcal{Z}_{\left(k\right)} $.
There exists subspace of representations (SoR) $ \mathcal{Z}'_k $ in the latent space $ \mathcal{Z}' $ with its data $ z' $, SoR $ \mathcal{Z}_k $ and $ \mathcal{V}_k $ in the latent space $ \mathcal{Z}_{\left(k\right)} $ with its data $ z_k $ and $ v_k $.
There are two mappings by $ \mathrm{g}:\mathcal{X}_k\mapsto \mathcal{Z}'_k $ and  $ \mathrm{d}_k:\mathcal{Z}'_k\mapsto \mathcal{Z}_k $.

\vspace{0.5em}
\noindent
The back-propagation of ACoRL has 3 processes:

\begin{enumerate}[label=(\roman*)]
    \item Projection model $ \mathrm{d}_k $ tries to modify SoR $ \mathcal{Z}_k $ to be highly coincident with $ \mathcal{V}_k $, where $ \mathcal{V}_k $ is the SoR for pre-trained model $ \mathrm{m}_k $.
          Then $ \mathcal{Z}_k $ can fully represent $ \mathcal{V}_k $.
    \item Assuming at this training state, the latent space where the representation of the Alliance Model can maximally reach in $ \mathcal{Z}' $ is denoted as $ \mathcal{Z}'_{\mathcal{A}} $.
          To ensure the mapping from $ \mathcal{Z}'_k $ to $ \mathcal{Z}_k $, the model $ \mathrm{d}_k $ will have a positive gradient $ \lambda \cdot \partial \mathcal{L} _{\mathrm{adv}}/\partial \theta _{\mathrm{d}} $.
          However, due to GRL, in the mapping from $ \mathcal{X}_k $ to $ \mathcal{Z}_k $, the gradient is inverted to be $ -\lambda \cdot \partial \mathcal{L} _{\mathrm{adv}}/\partial \theta _{\mathrm{g}} $.
          Since $ \mathcal{Z}'_{\mathcal{A}} $ originally had a negative gradient, it is now positive.
          Similarly, $ \mathcal{Z}'_{k} $ becomes negative.
          This indicates that model $ \mathrm{g} $ will actively search for more complementary representations in the latent space under ACoRL.
          Due to the SoR of the newly trained model $ \mathcal{Z}'_{k+1} $ is highly consistent with the original $ \mathcal{Z}'_{k} $, and yet ACoRL can greatly enlarge this SoR.
          According to Explain 2, since $ \mathcal{Z}'_{\mathcal{A}}\cup\mathcal{Z}'_k \gg \mathcal{Z}'_{k+1}\cup\mathcal{Z}'_k $, the ACoRL is theoretically proved to have much superior performance.
    \item The task branch has a mapping by $ \mathrm{h}:\mathcal{Z}'\mapsto \mathcal{Y} $ to the task space $ \mathcal{Y} $ by alternative SoR of $ \mathcal{Z}'_{\mathcal{A}} $.
\end{enumerate}

\section{Experiment}
\label{sec:experiment}

\subsection{Experimental Setup}
% ACoRL will be experimented on two tasks.
\noindent
\subsubsection*{Experiment 1 (image classification; IC)}
The task is trained and tested on ImageNet-100 \cite{deng2009imagenet} train and validation set.
Function $ \ell_t $ is softmax for classification.
The models are ResNet-34 (A), MobileNet-V2 (B) and ResNet-18 (C).
The test metrics are reported using top-1 accuracy (Acc) in percentage (\%).

\vspace{0.2em}
\noindent
\subsubsection*{Experiment 2 (speaker verification; SV)}
This task is trained on VoxCeleb1+2 \cite{Nagrani19} and tested on VoxCeleb1-O.
Function $ \ell_t $ is AAM-Softmax \cite{Deng2018ArcFaceAA} for speaker embedding (SE) training.
The models are ECAPA-TDNN \cite{DesplanquesTD20} (A), x-vector \cite{Snyder2018XVectorsRD} (B) and ResNet-34 \cite{he2016deep} (C).
Models A, B, and C use log mel-spectrograms with 80, 24, and 80 dimensions, respectively.
The test metrics are presented using the equal error rate (EER) in percentage (\%).

\vspace{0.2em}
\noindent
\subsubsection*{Training Details}
For IC, the second last layer features are flattened and used as the representation.
For SV, the SE is used as the representation.
The AAM-Softmax margin is 0.2 with the scale of 32.
We consider two fusion methods -- late fusion (L.F.) and output fusion (O.F.).
For L.F., the MMF is implemented by concatenating the representations and passing them through a 3-layer MLP with 512 nodes in the hidden layer.
For O.F., the IC task uses a weighted sum of classifier outputs while the SV task uses logistic regression on the score outputs.
Similar to the knowledge distillation framework \cite{Hinton2015DistillingTK} for knowledge learning, the ACoRL task function $ \ell_{\mathrm{adv}} $ is implemented with Kullback-Leibler Divergence (KL) as the objective.
% The ACoRL task function $ \ell_{\mathrm{adv}} $ is implemented with mean square error (MSE) as the objective.
The value of the hyperparameter $ \lambda=1 $.
As the ACoRL training is performed sequentially, model B is trained on top of model A, and model C on top of models A and B
We then evaluate the fusion of A+B and A+B+C models.

\subsection{Evaluation Results}

\noindent
\subsubsection*{Experiment 1}
Four conclusions can be drawn from the Table \ref{tab:ic}:
\begin{enumerate}[label=(\roman*)]
    \item Although newly trained individual models exhibit slightly better performance than their ACoRL-trained counterparts, fusing ACoRL-trained models leads to much superior overall performance.
          We hypothesize that each individually trained model may over-focus on similar, redundant features, whereas ACoRL diversifies this focus.
    \item L.F. outperforms the O.F.
          In our analysis, in O.F., only a single weight per output is excessively less accurate when fusing diverse models.
          In contrast, L.F. more carefully integrates and learns from more primitive information, thereby leveraging complementary information to improve the system performance.
    \item Incrementally aggregating models can improve performance, but the gains diminish as more models are added.
          Nevertheless, the ACoRL framework exhibits considerably greater performance improvement over traditional MMF.
    \item Interestingly, the fusion of 2 models A+B under ACoRL is even better than that of A+B+C under MMF.
          This indicates that ACoRL is more efficient in MMF, i.e., fusing only a few models with a small amount of computation makes a big difference.

\end{enumerate}

\begin{table}[t]
    \centering
    % \footnotesize
    \setlength{\tabcolsep}{8.9pt}
    \renewcommand{\arraystretch}{1.0}
    \begin{tabular}{@{}lrccccc@{}}
        \toprule
        \multicolumn{2}{c}{Top-1 Acc(\%)} & A     & B                      & C                      & A+B                    & A+B+C                           \\ \midrule
        \multirow{2}{*}{MMF}              & L. F. & \multirow{4}{*}{80.62} & \multirow{2}{*}{81.74} & \multirow{2}{*}{77.54} & 82.40          & 82.76          \\
                                          & O. F. &                        &                        &                        & 81.98          & 82.26          \\ \cmidrule(r){1-2} \cmidrule(l){4-7}
        \multirow{2}{*}{+ACoRL}           & L. F. &                        & \multirow{2}{*}{79.62} & \multirow{2}{*}{75.82} & \textbf{83.94} & \textbf{84.22} \\
                                          & O. F. &                        &                        &                        & 81.18          & 81.80          \\ \bottomrule
    \end{tabular}
    \caption{Overall comparison on the image classification task.}
    \label{tab:ic}
    % \vspace{1em}
\end{table}

\begin{table}[t]
    \centering
    % \footnotesize
    \setlength{\tabcolsep}{10.1pt}
    \renewcommand{\arraystretch}{1.0}
    \begin{tabular}{@{}lrccccc@{}}
        \toprule
        \multicolumn{2}{c}{EER(\%)} & A     & B                     & C                     & A+B                   & A+B+C                         \\ \midrule
        \multirow{2}{*}{MMF}        & L. F. & \multirow{4}{*}{0.90} & \multirow{2}{*}{2.97} & \multirow{2}{*}{1.07} & 0.84          & 0.81          \\                                & O. F. &                       &                       &                       & 0.87  & 0.86 \\ \cmidrule(r){1-2} \cmidrule(l){4-7}
        \multirow{2}{*}{+ACoRL}     & L. F. &                       & \multirow{2}{*}{3.12} & \multirow{2}{*}{1.24} & \textbf{0.78} & \textbf{0.73} \\
                                    & O. F. &                       &                       &                       & 0.80          & 0.77          \\ \bottomrule
    \end{tabular}
    \caption{Overall comparison on speaker verification task.}
    \label{tab:sv}
    % \vspace{1em}
\end{table}

\vspace{0.5em}
\noindent
\subsubsection*{Experiment 2}
Table \ref{tab:ic} could also show the 4 conclusions from experiment 1.
However, the performance gains conferred by L.F. over O.F. are limited for SV, compared to IC.
This may be attributed to two factors:
\begin{enumerate}[label=(\roman*)]
    \item SV in this experiment employs cosine similarity for scoring, which is a parameter-free process.
          Thus, the information contained in the SV scoring results is highly consistent with the SEs.
    \item The strong correlation between SV scores and SEs limits the potential for L.F. to outperform over O.F.
          Moreover, L.F. requires more sophisticated training and inference procedures than O.F., with greater computational cost.
          This is why SV typically uses O.F. instead of L.F. in MMF.
          In comparison, in IC, there is a complex mapping relationship between representations and output predictions.
          Generally, lower-level representations generally encapsulate richer information.
          Therefore, L.F. can improve performance more notably than O.F. for IC.
\end{enumerate}

\begin{figure}[t]
    \centering
    \includegraphics[width=0.45\textwidth]{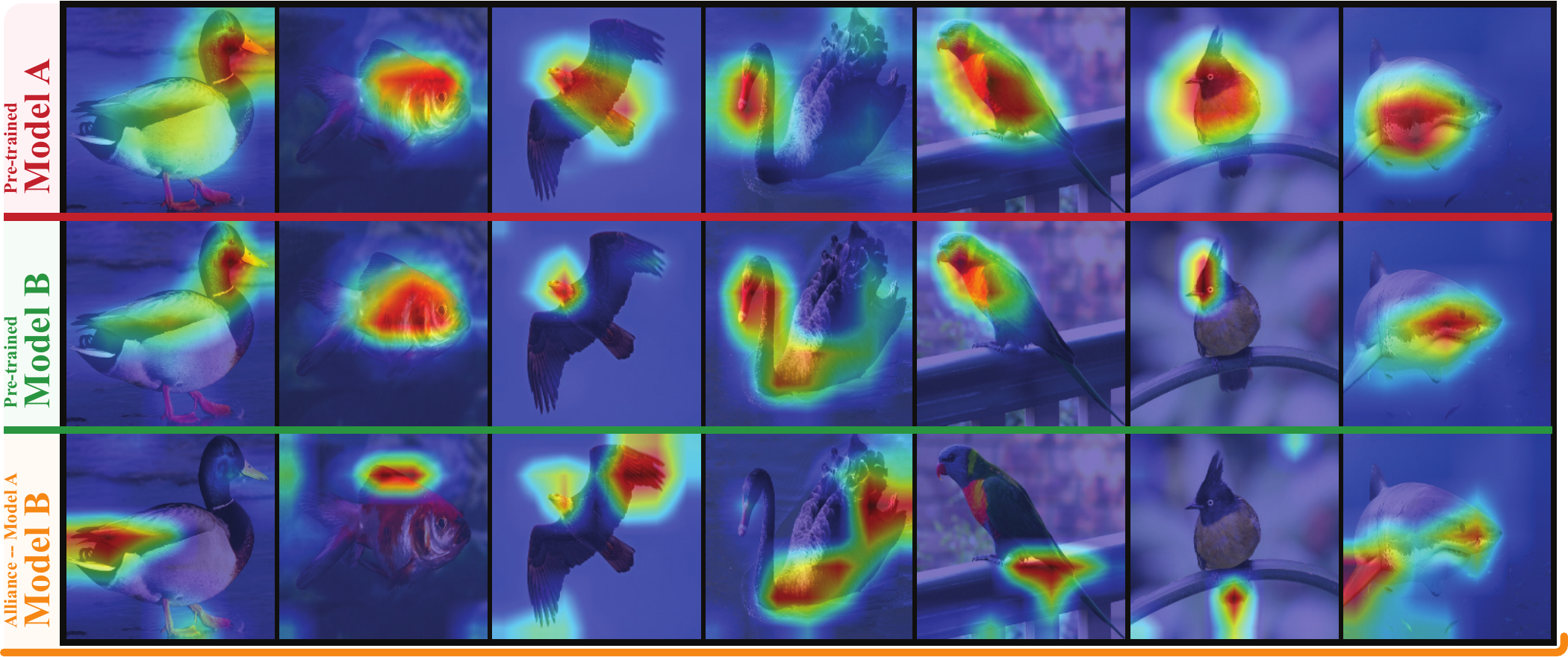}
    \caption{Attribution analysis of image samples on ImageNet-100 by GradCAM \cite{Chattopadhyay2017GradCAMGG} method.
        From top to bottom are the newly trained model A, model B, and model B trained under ACoRL on pre-trained model A.
        Red regions represent focus.}
    \label{fig:attri_ic}
    % \vspace{1em}
\end{figure}

\begin{figure}[t]
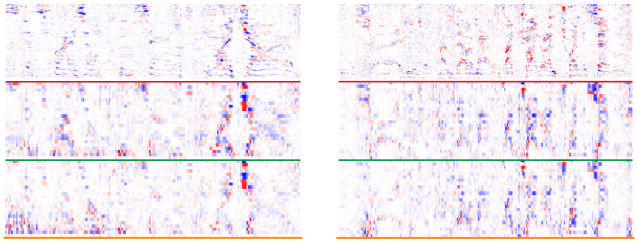

    \centering
    \FigSpect{attri_sv_0}
    \FigSpect{attri_sv_1}
    \caption{Attribution analysis of audio samples on VoxCeleb by integrated gradients \cite{Sundararajan2017AxiomaticAF} method.
        Three rows are the same as Figure \ref{fig:attri_ic}.
        The blue and red regions represent positive and negative gradients.
        \textcolor[rgb]{0.6,0.6,0.6}{(\textit{click on the figure to hear the sound})}}
    \label{fig:attri_sv}
    % \vspace{1em}
\end{figure}

\vspace{0.5em}
\noindent
\subsubsection*{Attribution Analysis}
The visualization in Figure \ref{fig:attri_ic} (with grad-CAM \cite{selvaraju2016grad} method) shows that before the addition of ACoRL, models A and B focused on similar regions.
ACoRL diminishes the current attention and increases focus on complementary, task-relevant aspects, implying that this newly trained alliance model leverages previously unfocused information to accomplish the task in an alternative way.
Similar to the attribution analysis of SV by J. Zhang et al. \cite{Zhang2023ASO}, Figure \ref{fig:attri_sv} (with integrated gradient \cite{sundararajan2017axiomatic} method) indicates SV focuses on a more fine-grained, abstract region, but reaches the same conclusion.
Compared with the second row (newly trained model B) in Figure \ref{fig:attri_sv}, the most notable regions on the third row (model B under ACoRL) are diminished, but instead, many other interesting small regions appear.
These newly added regions contain complementary knowledge, which is essential for improving the performance of the MMF.

\vspace{0.5em}
\section{Conclusions}
\label{sec:conclusion}

This paper proposed an adversarial complementary representation learning framework (ACoRL) that enables component models to learn maximally distinct but complementary representations, which is further confirmed by later attribution analysis.
Both explains and experimental results demonstrate that our method is more efficient and outperforms conventional methods.
ACoRL provides a generalizable method for improving the efficiency and robustness of multi-model fusion (MMF), offering a new avenue for future research.

\section{Acknowledgement}

Supported by the Key Research and Development Program of Guangdong Province (grant No. 2021B0101400003) and the Corresponding author is Zuheng Kang (bob.kang@qq.com).

% \clearpage
\bibliographystyle{IEEEbib}
\bibliography{refs}

\begin{thebibliography}{10}

\bibitem{huang2020fusion}
Shih-Cheng Huang, Anuj Pareek, Saeed Seyyedi, Imon Banerjee, and Matthew~P
  Lungren,
\newblock ``Fusion of medical imaging and electronic health records using deep
  learning: a systematic review and implementation guidelines,''
\newblock {\em NPJ digital medicine}, vol. 3, no. 1, pp. 136, 2020.

\bibitem{Kang2022SpeechEQSE}
Zuheng Kang, Junqing Peng, Jianzong Wang, and Jing Xiao,
\newblock ``Speecheq: Speech emotion recognition based on multi-scale unified
  datasets and multitask learning,''
\newblock in {\em Conference of the International Speech Communication
  Association (INTERSPEECH)}, 2022.

\bibitem{Lin2022ANM}
Shan Lin, Xingzhong Nong, Jianqiang Luo, and Chenen Wang,
\newblock ``A novel multi-model stacking ensemble learning method for metro
  traction energy prediction,''
\newblock {\em IEEE Access}, vol. 10, pp. 129231--129244, 2022.

\bibitem{Gao2019MultimodelFM}
Xizhan Gao, Quansen Sun, Haitao Xu, Dong Wei, and Jianqiang Gao,
\newblock ``Multi-model fusion metric learning for image set classification,''
\newblock {\em Knowledge Based System}, vol. 164, pp. 253--264, 2019.

\bibitem{peng2021effective}
Junyi Peng, Xiaoyang Qu, Rongzhi Gu, Jianzong Wang, Jing Xiao, Lukas Burget,
  and Jan Cernocky,
\newblock ``Effective phase encoding for end-to-end speaker verification.,''
\newblock in {\em Conference of the International Speech Communication
  Association (INTERSPEECH)}, 2021, pp. 2366--2370.

\bibitem{xu2020social}
Guangxia Xu, Weifeng Li, and Jun Liu,
\newblock ``A social emotion classification approach using multi-model
  fusion,''
\newblock {\em Future Generation Computer Systems}, vol. 102, pp. 347--356,
  2020.

\bibitem{polinati2019review}
Srinivasu Polinati and Ravindra Dhuli,
\newblock ``A review on multi-model medical image fusion,''
\newblock in {\em International Conference on Communication and Signal
  Processing (ICCSP)}. IEEE, 2019, pp. 0554--0558.

\bibitem{lin2022multi}
Mingqiang Lin, Denggao Wu, Jinhao Meng, Ji~Wu, and Haitao Wu,
\newblock ``A multi-feature-based multi-model fusion method for state of health
  estimation of lithium-ion batteries,''
\newblock {\em Journal of Power Sources}, vol. 518, pp. 230774, 2022.

\bibitem{tian2021multi}
Zhongda Tian and Hao Chen,
\newblock ``Multi-step short-term wind speed prediction based on integrated
  multi-model fusion,''
\newblock {\em Applied Energy}, vol. 298, pp. 117248, 2021.

\bibitem{Li2019AdaptingIS}
C.~Li, Dongliang He, Xiao Liu, Yukang Ding, and Shilei Wen,
\newblock ``Adapting image super-resolution state-of-the-arts and learning
  multi-model ensemble for video super-resolution,''
\newblock {\em Conference on Computer Vision and Pattern Recognition (CVPR)},
  pp. 2033--2040, 2019.

\bibitem{deb2022multi}
Sagar~Deep Deb, Rajib~Kumar Jha, Kamlesh Jha, and Prem~S Tripathi,
\newblock ``A multi model ensemble based deep convolution neural network
  structure for detection of covid19,''
\newblock {\em Biomedical signal processing and control}, vol. 71, pp. 103126,
  2022.

\bibitem{gao2019multi}
Xizhan Gao, Quansen Sun, Haitao Xu, Dong Wei, and Jianqiang Gao,
\newblock ``Multi-model fusion metric learning for image set classification,''
\newblock {\em Knowledge-Based Systems}, vol. 164, pp. 253--264, 2019.

\bibitem{Deng2019DeepMF}
Zijun Deng, Lei Zhu, Xiaowei Hu, Chi-Wing Fu, Xuemiao Xu, Qing Zhang, Jing Qin,
  and Pheng-Ann Heng,
\newblock ``Deep multi-model fusion for single-image dehazing,''
\newblock {\em International Conference on Computer Vision (ICCV)}, pp.
  2453--2462, 2019.

\bibitem{Jiang2020ATMFNAM}
Kui Jiang, Zhongyuan Wang, Peng Yi, Guangcheng Wang, Ke~Gu, and Junjun Jiang,
\newblock ``Atmfn: Adaptive-threshold-based multi-model fusion network for
  compressed face hallucination,''
\newblock {\em IEEE Transactions on Multimedia}, vol. 22, pp. 2734--2747, 2020.

\bibitem{zhang2020feature}
Ying Zhang, Rongrong Zhang, Qunfei Ma, Yanhao Wang, Qingqing Wang, Zihao Huang,
  and Linyan Huang,
\newblock ``A feature selection and multi-model fusion-based approach of
  predicting air quality,''
\newblock {\em ISA transactions}, vol. 100, pp. 210--220, 2020.

\bibitem{huh2023voxsrc}
Jaesung Huh, Andrew Brown, Jee-weon Jung, Joon~Son Chung, Arsha Nagrani, Daniel
  Garcia-Romero, and Andrew Zisserman,
\newblock ``Voxsrc 2022: The fourth voxceleb speaker recognition challenge,''
\newblock {\em arXiv preprint arXiv:2302.10248}, 2023.

\bibitem{sadjadi20222021}
Seyed~Omid Sadjadi, Craig Greenberg, Elliot Singer, Lisa Mason, and Douglas
  Reynolds,
\newblock ``The 2021 nist speaker recognition evaluation,''
\newblock {\em arXiv preprint arXiv:2204.10242}, 2022.

\bibitem{zhao2021speakin}
Miao Zhao, Yufeng Ma, Min Liu, and Minqiang Xu,
\newblock ``The speakin system for voxceleb speaker recognition challange
  2021,''
\newblock {\em arXiv preprint arXiv:2109.01989}, 2021.

\bibitem{yi2023audio}
Jiangyan Yi, Chenglong Wang, Jianhua Tao, Xiaohui Zhang, Chu~Yuan Zhang, and
  Yan Zhao,
\newblock ``Audio deepfake detection: A survey,''
\newblock {\em arXiv preprint arXiv:2308.14970}, 2023.

\bibitem{Liao2020EarlyBO}
Yixin Liao, Yuxuan Peng, Songlin Shi, Victor~Guang Shi, and Xiaohong Yu,
\newblock ``Early box office prediction in china's film market based on a
  stacking fusion model,''
\newblock {\em Annals of Operations Research}, vol. 308, pp. 321 -- 338, 2020.

\bibitem{Shen2018MEALME}
Zhiqiang Shen, Zhankui He, and X.~Xue,
\newblock ``Meal: Multi-model ensemble via adversarial learning,''
\newblock {\em Association for the Advancement of Artificial Intelligence
  (AAAI)}, 2019.

\bibitem{Nam2021DiversityMW}
Gi~Cheon Nam, Jongmin Yoon, Yoonho Lee, and Juho Lee,
\newblock ``Diversity matters when learning from ensembles,''
\newblock in {\em Neural Information Processing Systems}, 2021.

\bibitem{Zhang2018AdversarialCL}
Xiaolin Zhang, Yunchao Wei, Jiashi Feng, Yi~Yang, and Thomas~S. Huang,
\newblock ``Adversarial complementary learning for weakly supervised object
  localization,''
\newblock {\em Conference on Computer Vision and Pattern Recognition (CVPR)},
  pp. 1325--1334, 2018.

\bibitem{cao2020ensemble}
Yue Cao, Thomas~Andrew Geddes, Jean Yee~Hwa Yang, and Pengyi Yang,
\newblock ``Ensemble deep learning in bioinformatics,''
\newblock {\em Nature Machine Intelligence}, vol. 2, no. 9, pp. 500--508, 2020.

\bibitem{ganaie2022ensemble}
Mudasir~A Ganaie, Minghui Hu, AK~Malik, M~Tanveer, and PN~Suganthan,
\newblock ``Ensemble deep learning: A review,''
\newblock {\em Engineering Applications of Artificial Intelligence}, vol. 115,
  pp. 105151, 2022.

\bibitem{dong2020survey}
Xibin Dong, Zhiwen Yu, Wenming Cao, Yifan Shi, and Qianli Ma,
\newblock ``A survey on ensemble learning,''
\newblock {\em Frontiers of Computer Science}, vol. 14, pp. 241--258, 2020.

\bibitem{mohammed2023comprehensive}
Ammar Mohammed and Rania Kora,
\newblock ``A comprehensive review on ensemble deep learning: Opportunities and
  challenges,''
\newblock {\em Journal of King Saud University-Computer and Information
  Sciences}, 2023.

\bibitem{zhou2023inverse}
Sanping Zhou, Jinjun Wang, Le~Wang, Xingyu Wan, Siqi Hui, and Nanning Zheng,
\newblock ``Inverse adversarial diversity learning for network ensemble,''
\newblock {\em IEEE Transactions on Neural Networks and Learning Systems},
  2023.

\bibitem{Ganin2014UnsupervisedDA}
Yaroslav Ganin and Victor~S. Lempitsky,
\newblock ``Unsupervised domain adaptation by backpropagation,''
\newblock {\em ArXiv}, vol. abs/1409.7495, 2014.

\bibitem{Huang2021WhatMM}
Yu~Huang, Chenzhuang Du, Zihui Xue, Xuanyao Chen, Hang Zhao, and Longbo Huang,
\newblock ``What makes multimodal learning better than single (provably),''
\newblock in {\em Neural Information Processing Systems}, 2021.

\bibitem{deng2009imagenet}
Jia Deng, Wei Dong, Richard Socher, Li-Jia Li, Kai Li, and Li~Fei-Fei,
\newblock ``Imagenet: A large-scale hierarchical image database,''
\newblock in {\em Conference on Computer Vision and Pattern Recognition
  (CVPR)}. IEEE, 2009, pp. 248--255.

\bibitem{Nagrani19}
Arsha Nagrani, Joon~Son Chung, Weidi Xie, and Andrew Zisserman,
\newblock ``Voxceleb: Large-scale speaker verification in the wild,''
\newblock {\em Computer Science and Language}, 2019.

\bibitem{Deng2018ArcFaceAA}
Jiankang Deng, J.~Guo, and Stefanos Zafeiriou,
\newblock ``Arcface: Additive angular margin loss for deep face recognition,''
\newblock {\em Conference on Computer Vision and Pattern Recognition (CVPR)},
  pp. 4685--4694, 2018.

\bibitem{DesplanquesTD20}
Brecht Desplanques, Jenthe Thienpondt, and Kris Demuynck,
\newblock ``{ECAPA-TDNN:} emphasized channel attention, propagation and
  aggregation in {TDNN} based speaker verification,''
\newblock in {\em Conference of the International Speech Communication
  Association (INTERSPEECH)}. 2020, pp. 3830--3834, {International Symposium on
  Computer Architecture (ISCA)}.

\bibitem{Snyder2018XVectorsRD}
David Snyder, Daniel Garcia-Romero, Gregory Sell, Daniel Povey, and Sanjeev
  Khudanpur,
\newblock ``X-vectors: Robust dnn embeddings for speaker recognition,''
\newblock {\em International Conference on Acoustics, Speech and Signal
  Processing (ICASSP)}, pp. 5329--5333, 2018.

\bibitem{he2016deep}
Kaiming He, Xiangyu Zhang, Shaoqing Ren, and Jian Sun,
\newblock ``Deep residual learning for image recognition,''
\newblock in {\em Proceedings of the IEEE conference on computer vision and
  pattern recognition}, 2016, pp. 770--778.

\bibitem{Hinton2015DistillingTK}
Geoffrey~E. Hinton, Oriol Vinyals, and Jeffrey Dean,
\newblock ``Distilling the knowledge in a neural network,''
\newblock {\em ArXiv}, vol. abs/1503.02531, 2015.

\bibitem{Chattopadhyay2017GradCAMGG}
Aditya Chattopadhyay, Anirban Sarkar, Prantik Howlader, and Vineeth~N.
  Balasubramanian,
\newblock ``Grad-cam++: Generalized gradient-based visual explanations for deep
  convolutional networks,''
\newblock {\em IEEE Workshop on Applications of Computer Vision (WACV)}, pp.
  839--847, 2017.

\bibitem{Sundararajan2017AxiomaticAF}
Mukund Sundararajan, Ankur Taly, and Qiqi Yan,
\newblock ``Axiomatic attribution for deep networks,''
\newblock in {\em International Conference on Machine Learning (ICML)}, 2017.

\bibitem{selvaraju2016grad}
Ramprasaath~R Selvaraju, Abhishek Das, Ramakrishna Vedantam, Michael Cogswell,
  Devi Parikh, and Dhruv Batra,
\newblock ``Grad-cam: Why did you say that?,''
\newblock {\em arXiv preprint arXiv:1611.07450}, 2016.

\bibitem{Zhang2023ASO}
Jian Zhang, Liang He, Xiao~Song Guo, and Jingrong Ma,
\newblock ``A study on visualization of voiceprint feature,''
\newblock {\em Conference of the International Speech Communication Association
  (INTERSPEECH)}, pp. 2233--2237, 2023.

\bibitem{sundararajan2017axiomatic}
Mukund Sundararajan, Ankur Taly, and Qiqi Yan,
\newblock ``Axiomatic attribution for deep networks,''
\newblock in {\em International conference on machine learning}. PMLR, 2017,
  pp. 3319--3328.

\end{thebibliography}

\end{document}